# Adversarial Patch Attacks on Monocular Depth Estimation Networks

**KOICHIRO YAMANAKA**[1], **RYUTAROH MATSUMOTO**[2,3], (Member, IEEE), **KEITA TAKAHASHI**[1], (Member, IEEE), AND **TOSHIAKI FUJII**[1], (Member, IEEE)

[1]Graduate School of Engineering, Nagoya University, Nagoya 464-8603, Japan
[2]Department of Information and Communications Engineering, Tokyo Institute of Technology, Tokyo 152-8550, Japan
[3]Department of Mathematical Sciences, Aalborg University, 9220 Aalborg, Denmark

Corresponding author: Koichiro Yamanaka (koichiro.ymnk@fujii.nuee.nagoya-u.ac.jp)

**ABSTRACT** Thanks to the excellent learning capability of deep convolutional neural networks (CNN), monocular depth estimation using CNNs has achieved great success in recent years. However, depth estimation from a monocular image alone is essentially an ill-posed problem, and thus, it seems that this approach would have inherent vulnerabilities. To reveal this limitation, we propose a method of adversarial patch attack on monocular depth estimation. More specifically, we generate artificial patterns (adversarial patches) that can fool the target methods into estimating an incorrect depth for the regions where the patterns are placed. Our method can be implemented in the real world by physically placing the printed patterns in real scenes. We also analyze the behavior of monocular depth estimation under attacks by visualizing the activation levels of the intermediate layers and the regions potentially affected by the adversarial attack.

## I. INTRODUCTION

Estimating pixel-wise depth from 2-D images has become increasingly important with the recent development of autonomous driving, augmented realities (AR), and robotics. A large body of previous work has been devoted to depth estimation from stereo or more than two images [1]–[4]. At the same time, monocular depth estimation [5]–[8], in which depth is estimated from a single image,[1] has attracted attention due to its less demanding hardware requirements. Monocular depth estimation has been greatly enhanced by the excellent learning capability of deep convolutional neural networks (CNN). As a result, current state-of-the-art results with monocular depth estimation are quite impressive, and seemingly comparable to those with stereo methods (see Fig. 1, where (a) is the input image and (b) is the depth estimated by Guo *et al.* [7]). However, monocular depth estimation is essentially an ill-posed problem because a monocular image alone does not contain sufficient physical cues for scene depth. Instead of using the physical cues, these methods seem to rely on implicit knowledge (e.g., the color, vertical position, or shadows) that are learned from the training dataset [12]. We argue that monocular depth estimation depends too much on non-depth features in the given image, which makes it quite vulnerable to attacks.

To reveal the limitation mentioned above, we propose a method of adversarial patch attack for CNN-based monocular depth estimation. Specifically, we generate artificial patterns (adversarial patches) that can fool the target methods into estimating an incorrect depth for the regions where the patterns are placed. Figure 1(c) shows an example of our adversarial patches superimposed on the input image. As shown in (d), Guo *et al.*'s method [7] failed to estimate correct depth in the region where the patch was located; closer depth values were obtained than the original result in (b), as was intended with our design for this pattern. In this case, the attack was conducted in a digital manner; we digitally manipulated the pixel values of the input image to superimpose the patch. Our method can also be implemented in the real world, and we have achieved similar effects by physically placing the printed patterns in a real scene.

Moreover, to further analyze the behavior of monocular depth estimation under attacks, we visualize the

The associate editor coordinating the review of this manuscript and approving it for publication was Syed Islam.

[1]Generally, monocular depth estimation includes techniques that use a temporal sequence of images captured from a single camera [9]–[11]. However, in this article, we focus on methods that use only a single image from a single viewpoint for depth estimation.

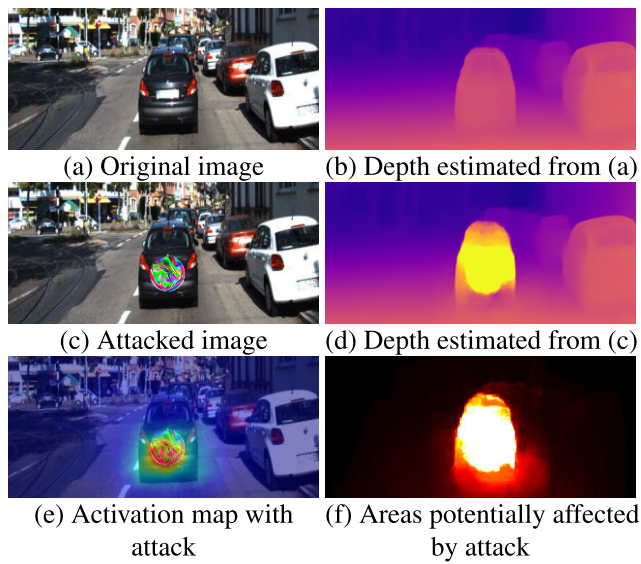

**FIGURE 1.** Monocular depth estimation [7] and our adversarial patch attack.

activation levels of the intermediate layers (Fig. 1(e)) and the regions that are potentially affected by adversarial attacks (Fig. 1(f)). These visualizations lead to a deeper understanding of the mechanism by which adversarial patches affect the target CNN. Our source code, learned patches and demo video are available at https://www.fujii.nuee.nagoya-u.ac.jp/Research/MonoDepth.

## II. BACKGROUND
### A. ADVERSARIAL ATTACKS
#### 1) DIGITAL ADVERSARIAL ATTACKS

Biggio *et al.* [13] were the first to demonstrate that deep neural networks (DNN) could be deceived by vicious attacks on the input. After that, a number of different adversarial attacks were proposed for image classification tasks [14]–[19]. Their purpose was to find small perturbations to be added to the original image to cause mis-classification. Compared to classification tasks, fewer works have been conducted for regression tasks. Hendrik Metzen *et al.* [20] demonstrated that nearly imperceptible perturbations could also fool an image segmentation method into producing incorrect results. Following their work, Xie *et al.* [21] proposed a method that can deceive both segmentation and object detection models simultaneously, and Wei *et al.* [22] extended the target of attack from an image to a video. More recently, Zhang *et al.* [23] attacked monocular depth estimation.

It should be noted that these methods implicitly assume "digital adversarial attack", in which the attacks are implemented by digitally manipulating the pixel values on the input image. The perturbation pattern is usually designed to have small amplitude, and thus, the difference between the original image and the attacked image is imperceptible to the human eye. However, the perturbation patterns usually cover the entire image, which makes it unsuitable to implement them in the real world.

#### 2) PHYSICAL ADVERSARIAL ATTACKS
A number of studies have also been conducted on "physical adversarial attacks" that can be implemented in the real world, e.g., by placing printed patterns in a target scene. These patterns are not necessarily designed to be imperceptible to the human eye depending on the applications [24]–[27].

Kurakin *et al.* [28] demonstrated that images with adversarial patterns for a classification task created in [15] would remain adversarial when they are printed and captured by cameras. Athalye *et al.* [29] extended this idea to 3D physical adversarial objects. Eykholt *et al.* [30] showed that stop signs can be misclassified if various stickers are placed on top of them. Their adversarial objects were designed to be indistinguishable to the human eye, similarly to the case with "digital adversarial attacks".

In a similar vein, Brown *et al.* [24] took a small designed patch (adversarial patch), which was clearly visible to the eye, and placed it in target scenes to induce errors in a classification task. Their patches were designed to be placed anywhere in an input image. Moreover, as their patches are independent of target scenes, they can be used for "physical adversarial attacks" without prior knowledge of lighting conditions, camera angles, or other objects in the target scene. This is not trivial, as a pattern located in the real world usually receives a series of transformations (e.g., geometric transform, digitization, and color gamut transform) before it is recorded in a digital image, which would invalidate the effect as an adversarial pattern [31]. Following [24], adversarial patches have been used in several tasks such as face recognition [25], object detection [26] and optical flow estimation [27]. Komkov and Petiushko [25] attacked face recognition by sticking an adversarial patch on a hat. Ranjan *et al.* [27] conducted an adversarial patch attack on optical flow that is, to our knowledge, the first work to apply adversarial patch attacks to a regression problem. Note that our method can also be located in the context of adversarial patch attack for a regression problem.

### B. MONOCULAR DEPTH ESTIMATION
Monocular depth estimation refers to the process of predicting pixel-wise depth from a single image. As a seminal work, Eigen *et al.* [32] proposed a multi-scale CNN architecture that can produce pixel-wise depth estimation from a single image. Unlike other previous works in single image depth estimation [33]–[38], their network did not rely on hand crafted features. Since then, significant improvements have been made by using techniques such as incorporation of strong scene priors for surface normal estimation [36], conditional random fields [39], conversion from a regression problem to a classification problem [40], and a quantized ordinal regression problem [6]. Lee *et al.* [8] achieved state-of-the-art result by introducing novel local planar guidance layers located at multiple stages in the decoding phase. All of the methods mentioned above are trained in a supervised manner, where ground-truth depth taken by RGB-D cameras or 3D laser scanners are required to train the networks.

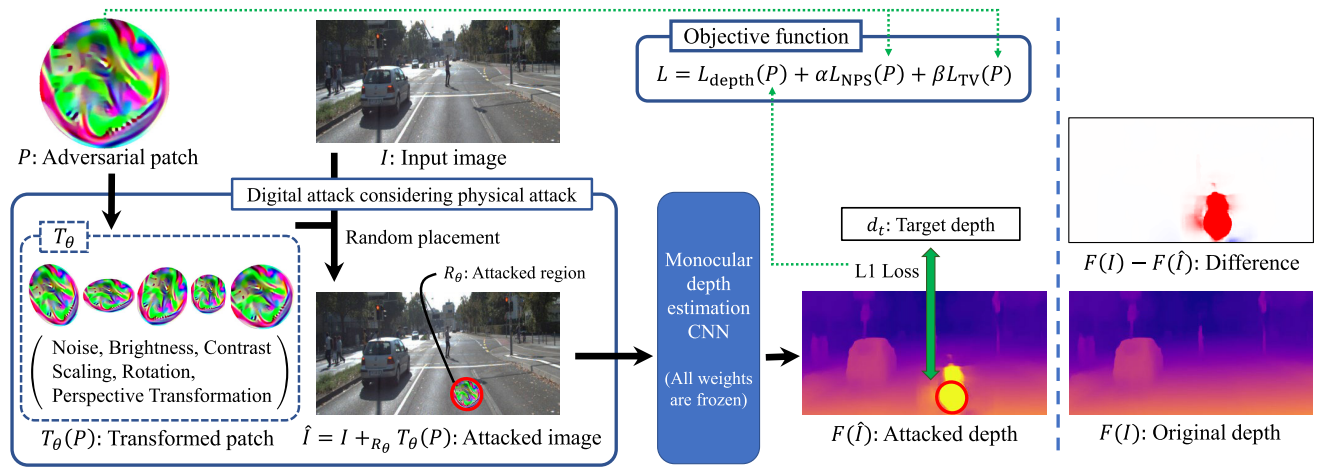

**FIGURE 2.** Overview of the proposed method.

Recently, semi-supervised and unsupervised methods have also been proposed. As examples of the semi-supervised approach, Chen et al. [41] used relative depth information and Kuznietsov et al. [42] used sparse depth data obtained from LiDAR. As for the unsupervised approaches, they require only rectified stereo image pairs to train networks. Xie et al. [43] proposed Deep3D to create a new right view from an input left image using depth image-based rendering [44], where a disparity map was estimated as an intermediate product. Garg et al. [45] extended this using a network similar to FlowNet [46]. However, since their network was not fully differentiable, they performed a Taylor series approximation to linearize their loss, which made the optimization difficult. To address this problem, Godard et al. [5] introduced differentiable bilinear sampling [47] into the framework of Xie et al. [43]. Additionally, they considered left-right consistency of the predicted disparities estimated from the given pair of stereo images. Their network consisted of an encoder and a decoder, where the encoder's structure was taken from VGG-16 [48] and the decoder was composed of stacked deconvolution layers and shortcut connections from the encoder. More recently, Guo et al. [7] adopted the concept of knowledge distillation [49], and trained their monocular depth estimation network to produce the same disparity maps as the ones that were predicted by a pre-trained stereo depth estimation network.

van Dijk and de Croon [12] analyzed the behavior of monocular depth estimation and argued that depth prediction would actually depend on non-depth features such as the vertical position and the texture.

### C. OUR CONTRIBUTION

We propose a method of adversarial patch attack for CNN-based monocular depth estimation methods. Similarly to some previous works [24]–[27], the patches used for our attack are recognizable to human eye. Zhang et al. [23] also attacked monocular depth estimation using imperceptibly small perturbation patterns covering the entire image, but their method was not applicable to the real world. In contrast, our patch-based approach enables physical attacks using printed patterns located in the target scene. Our method is similar to Ranjan et al.'s [27], which attacks optical flow CNNs using printed patches that are physically located in real scenes. We would like to stress several differences between our work and Ranjan et al.'s [27]. First, our method is designed to cause depth errors only around the patches rather than the entire image frame. Moreover, our method considers perspective transform [30] in the image formation model to increase the robustness of the attack in the real world. Finally, our target is monocular depth estimation, which we feel has inherent vulnerabilities due to its over-dependence on non-depth cues. To the best of our knowledge, we are the first to attack monocular depth estimation in real scenes. We also visualize the behavior of monocular depth CNNs under attack, which will contribute to a deeper understanding of monocular depth estimation in concert with other approaches (e.g., van Dijk and de Croon's [12]).

## III. PROPOSED METHOD
### A. OVERVIEW

Figure 2 illustrates the overview of our method. Given a target monocular depth estimation method implemented as a CNN, our goal is to derive an adversarial patch (denoted as $P$) that induces the target method to produce incorrect depth estimates for the region on the input image where the patch is located.

Our method is designed to be implemented in the real world; namely, we want to deceive the target method by physically placing printed patches in the target scene. To achieve this goal, we need to consider various shooting conditions in physical settings. When a patch is printed and placed in the target scene, it is subject to a series of transformations (luminance change, geometric transformations, noise, etc.) depending on the shooting condition before it is finally recorded in a digital image. Therefore, the imaging process of the patch is modeled so as to cover various shooting conditions.

Moreover, we implement the imaging process as fully differentiable so that the gradient with respect to the estimated depth can be propagated back to the patch through the network. During the training stage, we keep the target method's network unchanged and update only the adversarial patch through the framework of back-propagation. We utilize the Adam optimizer [50] for this purpose. We could also utilize custom-made iterative updating rules such as FGSM [15], but we found that using the Adam optimizer led to better results.

### B. MODEL OF IMAGING PROCESS

Let $F$ be the target monocular depth estimation CNN and $I$ the original input image. The estimated depth map is represented as $D = F(I)$. The adversarial patch we aim to derive is denoted as $P$. We assume that the patch $P$ receives various transformations $T_\theta$ through the imaging process and finally falls into the region $R_\theta$ in the input image. The attacked image is represented as $\hat{I} = I +_{R_\theta} T_\theta(P)$, where $+_{R_\theta}$ refers to the pixel overwriting on the region $R_\theta$.

We aim to train the patch $P$ so that the depth values in the region $R_\theta$ become a certain depth value $d_t$, regardless of the actual depth. This is formalized as

$$\underset{P}{\arg\min} \sum_{(i,j)\in R_\theta} \left| d_t - F_{i,j}(I +_{R_\theta} T_\theta(P)) \right|, \quad (1)$$

where $(i, j)$ denotes a pixel coordinate in the estimated depth map. By minimizing Eq. (1), we force the estimated depth in the region $R_\theta$ to be a specific depth value $d_t$. Depending on the value of $d_t$, the estimated depth is guided to be different from the actual depth.

The patch $P$ should be robust to various transformations in the imaging process. Therefore, we randomly change the transformation $T_\theta$ for each mini-batch during training. Specifically, $T_\theta$ includes random brightness shifts (in the $[-0.05, 0.05]$ range), contrast shifts (in the $[0.9, 1.1]$ range), addition of noise (in the $[-0.1, 0.1]$ range), scaling (with factors of $[0.1225, 0.2025]$), and rotation (in $[-20, 20]$ degrees). Finally, the patch $P$ receives perspective transformation, in which four vertices of the unit square $((0, 0), (0, 1), (1, 0), (1, 1))$ are randomly perturbed as

$$(v_x, v_y) \mapsto (v_x + u, v_y + v) \ (u, v \in [-0.1, 0.1]), \quad (2)$$

where $(v_x, v_y)$ represents each vertex of the unit square and $u, v$ represents horizontal and vertical shifts, respectively. Only a single patch is used to cover different spatial resolutions, since we include a larger range of scaling than [27] in the transform (in [27], several patches were learned for different resolutions). We used a sufficiently large resolution for the patch ($256 \times 256$ pixels) to reduce the unexpected effects of pixel interpolation when printing it.

### C. LOSS FUNCTION

Our loss function is defined for the target adversarial patch $P$ and is composed of three terms,

$$L = L_{\text{depth}}(P) + \alpha L_{\text{NPS}}(P) + \beta L_{\text{TV}}(P), \quad (3)$$

where $\alpha$ and $\beta$ are weighting coefficients that are determined experimentally.

#### 1) DEPTH LOSS

The most important term in our loss function, the depth loss, $L_{\text{depth}}$, is given in accordance with Eq. (1).

$$L_{\text{depth}}(P) = \sum_{(i,j)\in R_\theta} \left| d_t - F_{i,j}(I +_{R_\theta} T_\theta(P)) \right|. \quad (4)$$

When we attack more than one monocular depth estimation methods simultaneously, we replace Eq. (4) with the ensemble depth loss $L_{\text{depth}}^{\text{ens}}$, as

$$L_{\text{depth}}^{\text{ens}} = \sum_k \sum_{(i,j)\in R_\theta} \left| d_t - F_{i,j}^k(I +_{R_\theta} T_\theta(P)) \right|, \quad (5)$$

where $F^k$ denotes the $k$-th network.

#### 2) NON-PRINTABILITY SCORE (NPS)

We included NPS [51] in our loss function to limit the color space of the patch within the printable color gamut, as

$$L_{\text{NPS}}(P) = \sum_{i,j} \min_{\vec{c}\in C} \left\| \vec{p}_{i,j} - \vec{c} \right\|_1, \quad (6)$$

where $\|.\|_1$ denotes the $L_1$ norm and $p_{i,j}$ denotes the color vector of a pixel $(i, j)$ in the patch $P$. We seek the closest color vector $c$ from the set of printable colors $C$. A smaller $L_{\text{NPS}}$ means better printability.

#### 3) TOTAL VARIATION (TV)

Non-smooth patches are more likely to be affected by aliasing artifacts when they are printed and captured by a camera. Therefore, we encourage the smoothness of the patch $P$ by using total variation (TV) loss, similarly to [52].

$$L_{\text{TV}}(P) = \sum_{i,j} \sqrt{(p_{i,j} - p_{i+1,j})^2 + (p_{i,j} - p_{i,j+1})^2}. \quad (7)$$

### D. IMPLEMENTATION

#### 1) TARGET METHODS

As the target of our proposed attack, we used two state-of-the-art monocular depth estimation methods [7], [8].

Guo et al.'s method [7] is trained in an unsupervised manner using knowledge distillation. Their network consists of an encoder and a decoder. The encoder part is implemented using VGG-16 [48] and the decoder part is composed of stacked deconvolution layers and skip connections. The output from the network is a disparity map, which is converted into the depth map using the relation:

$$\text{disparity} = (\text{baseline} * \text{focal length})/\text{depth}, \quad (8)$$

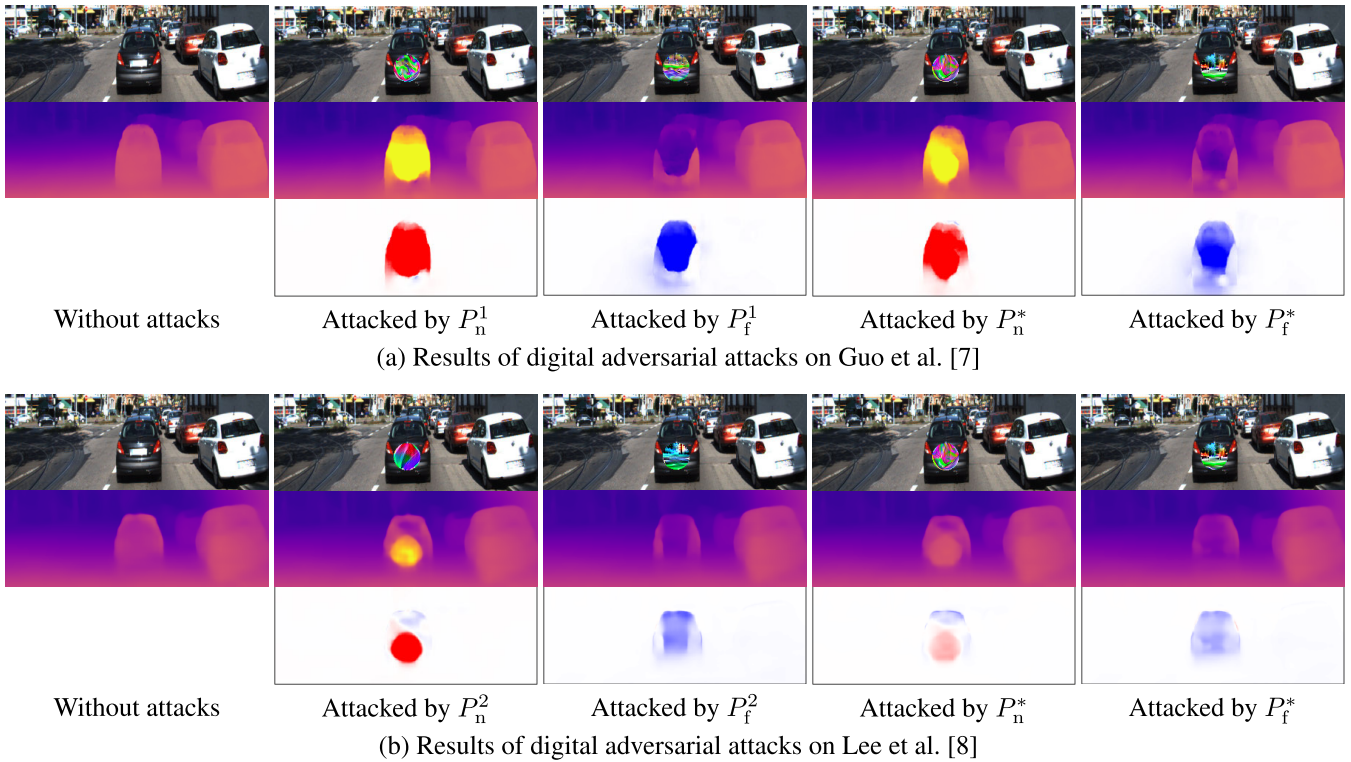

FIGURE 3. Results of digital adversarial attacks.

where the baseline and the focal length were provided in the KITTI dataset.

Meanwhile, Lee et al.'s [8] takes a supervised training framework, where the estimated depth is directly supervised by the corresponding ground-truth. Their network consists of an encoder and a decoder as well, but the structures are more complex than Guo et al.'s [7]; DenseNet-161 [53] was adopted as the encoder, and atrous spatial pyramid pooling [54] and local planar guidance layers were used in the decoder. This method is one of the top performing monocular depth estimation methods in the KITTI benchmark [55].

These two networks were pre-trained on the KITTI dataset [56] using a data-splitting rule proposed by Eigen et al. [32]. The Eigen split consisted of 22,600 stereo image pairs for training, 888 for validation, and 697 for testing.

#### 2) TRAINING DETAILS

We trained adversarial patches under several conditions. We tested two target depths ($d_t = 3$ m and 150 m) and three target configurations (individual and simultaneous attacks to either and both of [7] and [8]).

For training of our adversarial patch, we used only the left view images from the Eigen's 22,600 training split [32]. The width and height of the input images were set to 512 and 256 pixels, respectively to fit the input size of the target networks. The input images were randomly augmented by horizontal flipping, zooming with a factor of [0.8, 1], gamma correction in the range of [0.8, 1.2], illumination change with a factor of [0.8, 1.2], and color jittering in the range of [0.95, 1.05]. The resolution for the patch was set to $256 \times 256$ pixels, but its apparent size in the input image was changed by the patch transformer $T_\theta$.

We used a Linux-based PC equipped with a NVIDIA Geforce GTX 1080 Ti. The networks were implemented using Python version 3.6.9 and PyTorch [57] version 1.1.0. We used the Adam optimizer [50] with learning rate $10^{-3}$, and batch size was set to 8. The number of epochs was 40.

The resulting adversarial patches are shown in Fig. 4.

| Target depth | Target methods | | |
|---|---|---|---|
| | Guo et al. [7] | Lee et al. [8] | Guo et al. [7], Lee et al. [8] |
| $d_t = 3$ [m] | $P_n^1$ | $P_n^2$ | $P_n^*$ |
| $d_t = 150$ [m] | $P_f^1$ | $P_f^2$ | $P_f^*$ |

FIGURE 4. Adversarial patches trained with various conditions.

### IV. EXPERIMENTAL RESULTS
#### A. DIGITAL ADVERSARIAL ATTACK
We implemented the digital attack by superimposing the patches shown in Fig. 4 onto input images. Figure 3 shows the input images (w/ and w/o the attack), estimated depth maps, and the difference between the original and attacked depth maps.

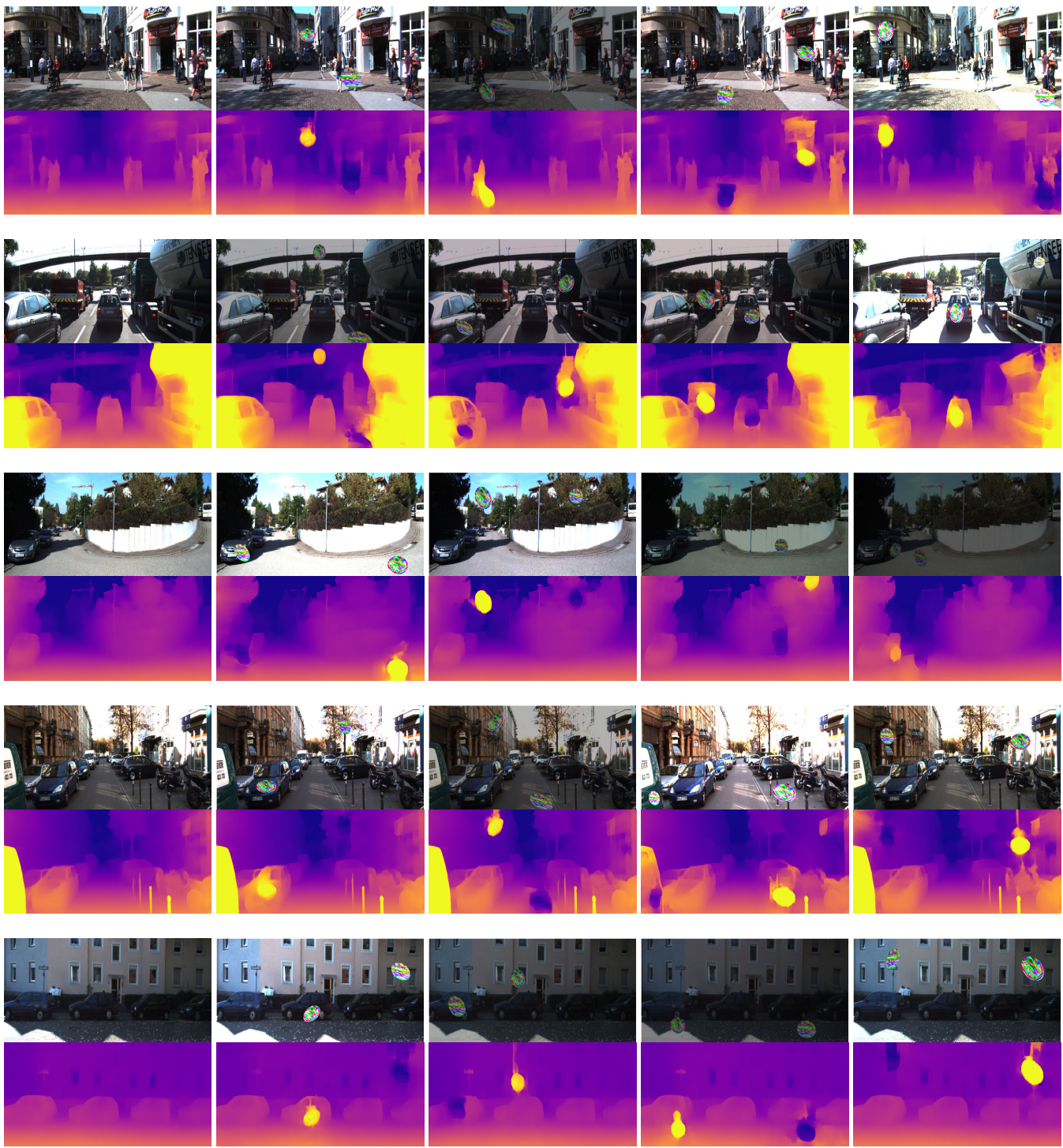

**FIGURE 5.** Results of digital adversarial attacks by $P_n^1$ and $P_f^1$ on Guo *et al.*'s method [7].

As shown in (a), our attack was quite effective for Guo *et al.*'s method [7]; in both cases, namely, where the patch was trained exclusively for Guo *et al.*'s method [7] and where it was trained for both the methods [7], [8], the network produced incorrect depths corresponding to $d_t$ in the regions where the patches were located.

In contrast, as shown in Fig. 3(b), the attack had a limited effect for Lee *et al.*'s method [8]; in particular, the patches trained for both methods were less effective. One possible reason for this is the higher complexity of Lee *et al.*'s method [8] compared to Guo *et al.*'s [7], which would bring more robustness to attacks. However, we conclude that our attack was effective to some extent because significant depth errors were induced by the adversarial patches.

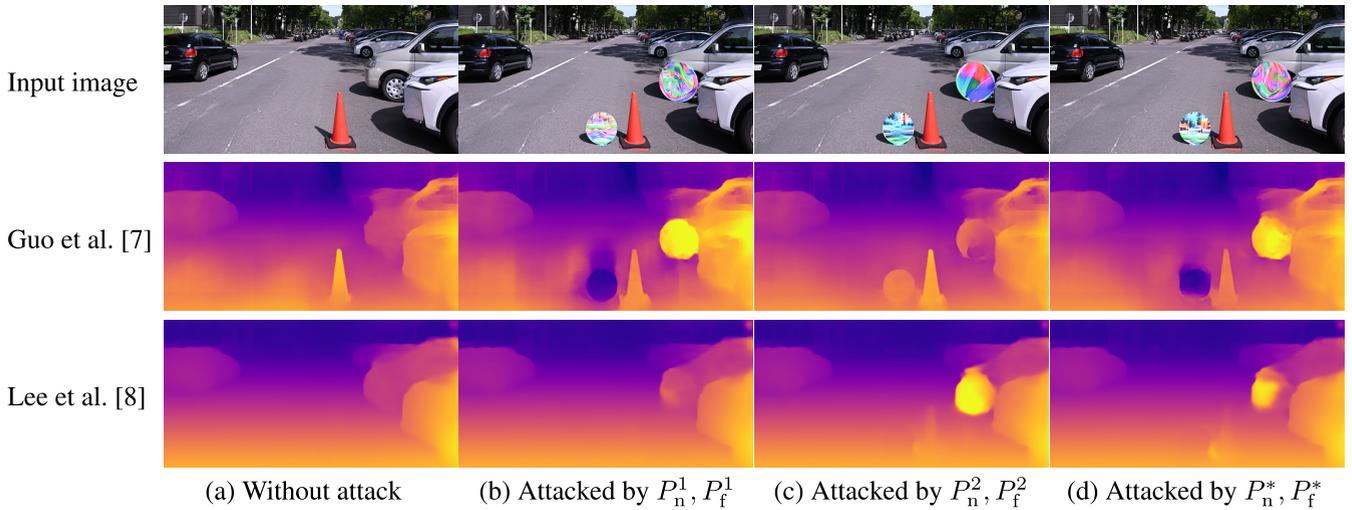

FIGURE 6. Results of physical adversarial attacks.

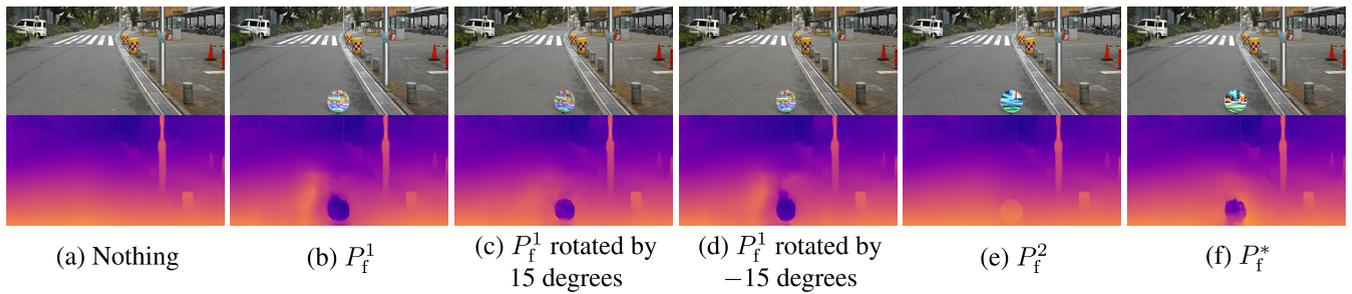

FIGURE 7. Results of physical adversarial attacks on Guo *et al.*'s method [7] under various conditions.

It should be noted that simultaneous attack was possible for both methods, which had different network structures and training strategies. See the results with $P_n^*$ and $P_f^*$ in Fig. 3. This would indicate the inherent vulnerability of monocular depth estimation, where few geometric cues are available from the input image itself.

To further demonstrate the effectiveness of our attack, we present additional results on Guo *et al.*'s method [7] in Fig. 5, which shows that our adversarial patches were effective in various conditions (location, scale, orientation, and background).

### B. PHYSICAL ADVERSARIAL ATTACK

We also conducted a physical adversarial attack by using the printed adversarial patches. Figure 6 shows several input images and estimated depth maps. We can see a similar tendency as the one with the digital adversarial attack: namely, the attack was less effective against Lee *et al.*'s method [8] than Guo *et al.*'s [7]. In particular, the effect of $P_f^2$ on Lee *et al.*'s method [8] was almost non-existent. This is seemingly related to the property of Lee *et al.*'s method that the estimated depth is closely related to the vertical position in the image; $P_f^2$ ($d_t = 150$ m) was placed near the bottom of the image, where Lee *et al.*'s method is more likely to produce small depth values. In contrast, when $P_f^2$ was placed near the top of the image, we obtained a result closer to our intention (see Fig.8). To conclude, both of the monocular depth estimation methods could be attacked in the real world, although the results were not always sufficient against Lee *et al.*'s method.

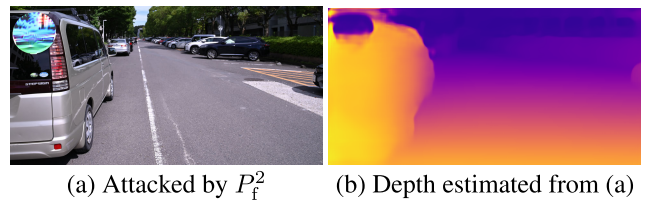

FIGURE 8. Successful case of our attack on Lee et al's method [8].

In Fig. 7, we present more results of the physical adversarial attack against Guo *et al.*'s method. In (b), the patch $P_f^1$ was located upright, but in (c) and (d), it was rotated by ± 15 degrees. The attack was effective for all three cases, indicating the robustness of the adversarial patch derived by our method. In (e), we can see that the patch $P_f^2$ (trained for Lee et al.'s method) had little effect on Guo et al.'s method, although the appearance of $P_f^2$ was similar to that of $P_f^*$, and in (f), $P_f^*$ was effective on the same method. This indicates that similarity to the human eye does not always correspond to the similarities to the depth estimation methods.

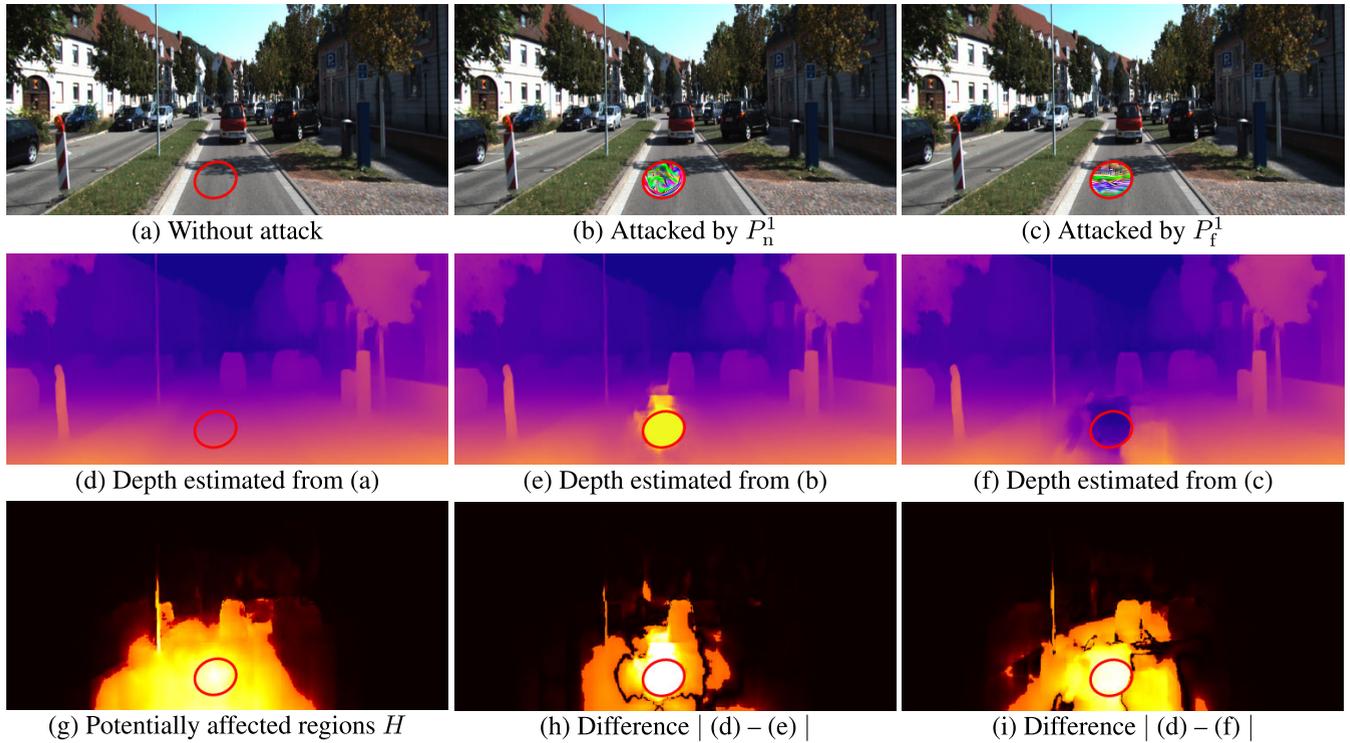

**FIGURE 9.** Results on visualizing potentially affected regions. Red lines indicate the region of $R_\theta$.

Please refer to the supplementary video for more results.

## V. ANALYSIS THROUGH VISUALIZATION

As discussed in the previous section, the effect of our adversarial attack was spatially localized in the resulting depth map; incorrect depth estimates were induced only in the regions around the adversarial patches. To analyze this effect, we present two methods for visualizing the effects of adversarial patches. First, we visualize the potential regions on a depth map where the depth values are likely to be affected by an adversarial patch. Second, we visualize the network's activation incurred by the adversarial patch, which helps to analyze the mechanism where the adversarial patch induces incorrect depth estimates. We use digital attack (adversarial patches superimposed by manipulating the image digitally) for this analysis, similarly to the work of van Dijk and de Croon [12]. We adopted Guo et al.'s method [7] as the target of visualization.

### A. POTENTIALLY AFFECTED REGIONS

Given a region $R_\theta$ for an adversarial patch to be placed in the input image $I$, we can predict the potential effect on the resulting depth value $D(u, v)$ for a pixel $(u, v)$ as

$$H(u, v) = \left| \sum_c \sum_{(i,j) \in R_\theta} \frac{\partial D(u, v)}{\partial I(i, j, c)} \right|, \quad (9)$$

where $D = F(I)$. In the above equation, the partial derivative can be obtained by standard back-propagation. The map $H(u, v)$ shows how much the estimated depth of each pixel $(u, v)$ is potentially affected by the adversarial patch located at $R_\theta$. Note here that we only consider the location $R_\theta$, and do not specify the pattern for the adversarial patch. Therefore, with this visualization, we actually analyze the intrinsic sensitivity of the pre-trained target network $F$.

Figure 9 shows an example of this visualization. The original input image and an attacked image are shown in (a), (b), and (c) from which Guo et al.'s method [7] predicted the depth maps shown in (d), (e), and (f), respectively. Shown in (g) is $H(u, v)$, a prediction of the potentially affected region, where $R_\theta$ was set to the region of the adversarial patch in (b) and (c). To verify this prediction, the difference between (d) and (e) is presented in (h) and the difference between (d) and (f) is presented in (i). As we can see, the predicted region in (g) was well aligned to the resulting depth differences in (h) and (i).

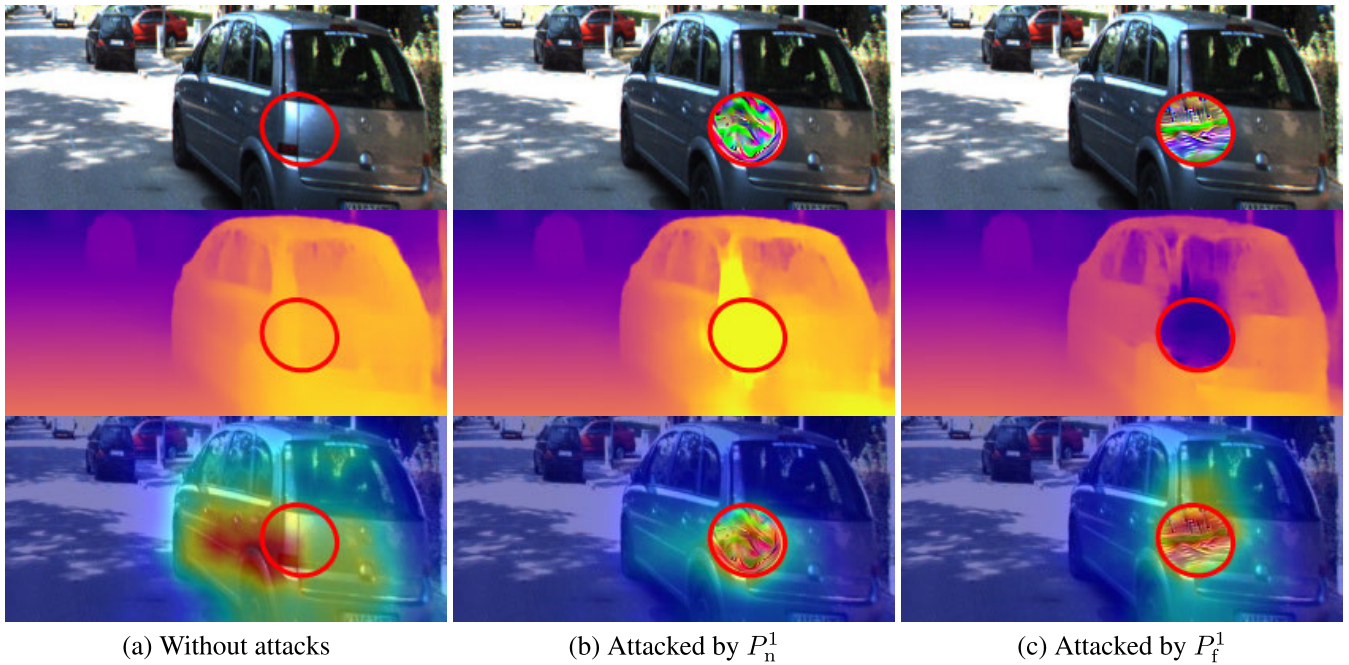

(a) Without attacks        (b) Attacked by $P_n^1$        (c) Attacked by $P_f^1$

**FIGURE 10.** Results on visualizing activation map of Guo's network [7]. Top row: input image. Middle row: estimated depth maps. Bottom row: visualization results. Red lines indicate the region of $R_\theta$.

### B. NETWORK ACTIVATION

For classification tasks, a well known method for visualizing network activation is Grad-CAM [58]. This method can be applied only for classification tasks, where the output from the network is typically a global label and no spatial information is involved. However, in our case, the output from the network is a pixel-wise depth map that depends on the location of a pixel. Therefore, we extend the idea of Grad-CAM to our problem as follows.

First, we focus on the encoder part of the target network $F$. We obtain a layer-wise activation map for a spatial position $(u, v)$ in the output depth map $D(u, v)$ as

$$G_m^{u,v}(i,j) = \sum_k \left| \frac{\partial D(u,v)}{\partial A_m^k(i,j)} \cdot A_m^k(i,j) \right|, \quad (10)$$

where $A_m^k(i, j)$ is the $k$-th feature map of the $m$-th convolution layer and $(i, j)$ denotes the spatial location on the feature map. We then resize activation maps $G_m^{u,v}(i,j)$ to the same size as the input image by bilinear interpolation and aggregate them over all the encoder's layers. We finally take the summation over the regions $(u, v) \in R_\theta$ to obtain the final activation map, as

$$G(i,j) = \sum_{(u,v) \in R_\theta} \left( \sum_m \text{Resize}(G_m^{u,v}(i,j)) \right). \quad (11)$$

This activation map shows the extent to which spatial neighbors are involved for the erroneous estimates on $(u, v) \in R_\theta$, when an adversarial patch covers the region $R_\theta$.

We present an example in Fig. 10, where input images, depth maps, and activation maps ($G(i, j)$) are shown for three cases: (a) without attacks, (b) attacked by $P_n^1$, and (c) attacked by $P_f^1$. In all cases, $R_\theta$ was set to the same region (the region of the adversarial patches). It should be noted that the activation heavily depended on the presence of the adversarial patches: in (a), the activation covered a wide area around $R_\theta$, but in (b) and (c), the activation seemingly concentrated on $R_\theta$. These results suggest that the adversarial patches attracted the attention of the target method to itself, which in turn led to incorrect depth estimation, as intended.

## VI. CONCLUSION

We have proposed a method of adversarial patch attack for CNN-based monocular depth estimation methods. The adversarial patches were trained to induce incorrect depth estimation around the region where they were located, in the presence of various transformations such as perspective transform, scaling, and translation. We demonstrated that our method can be implemented in the real world by placing a printed pattern in the target scene. We also analyzed the behavior of monocular depth methods under attack by visualizing the potentially affected regions and activation maps. To the best of our knowledge, we are the first to achieve physical adversarial attacks on depth estimation methods. Our future work will include extending our method to similar tasks such as optical flow estimation and stereo depth estimation. We hope our work will lead to a wider recognition of the vulnerability of monocular depth estimation methods and thus to the development of safer depth estimation techniques.

### REFERENCES

[1] J. L. Schonberger and J.-M. Frahm, "Structure-from-Motion revisited," in *Proc. IEEE Conf. Comput. Vis. Pattern Recognit. (CVPR)*, Jun. 2016, pp. 4104–4113.

[2] J. L. Schönberger, E. Zheng, J.-M. Frahm, and M. Pollefeys, "Pixelwise view selection for unstructured multi-view stereo," in *Proc. Eur. Conf. Comput. Vis. (ECCV)*, 2016, pp. 501–518.

[3] T.-C. Wang, A. A. Efros, and R. Ramamoorthi, "Depth estimation with occlusion modeling using light-field cameras," *IEEE Trans. Pattern Anal. Mach. Intell.*, vol. 38, no. 11, pp. 2170–2181, Nov. 2016.

[4] J.-R. Chang and Y.-S. Chen, "Pyramid stereo matching network," in *Proc. IEEE/CVF Conf. Comput. Vis. Pattern Recognit.*, Jun. 2018, pp. 5410–5418.

[5] C. Godard, O. M. Aodha, and G. J. Brostow, "Unsupervised monocular depth estimation with left-right consistency," in *Proc. IEEE Conf. Comput. Vis. Pattern Recognit. (CVPR)*, Jul. 2017, pp. 270–279.

[6] H. Fu, M. Gong, C. Wang, K. Batmanghelich, and D. Tao, "Deep ordinal regression network for monocular depth estimation," in *Proc. IEEE/CVF Conf. Comput. Vis. Pattern Recognit.*, Jun. 2018, pp. 2002–2011.

[7] X. Guo, H. Li, S. Yi, J. Ren, and X. Wang, "Learning monocular depth by distilling cross-domain stereo networks," in *Proc. Eur. Conf. Comput. Vis. (ECCV)*, 2018, pp. 484–500.

[8] J. Han Lee, M.-K. Han, D. Wook Ko, and I. Hong Suh, "From big to small: Multi-scale local planar guidance for monocular depth estimation," 2019, *arXiv:1907.10326*. [Online]. Available: http://arxiv.org/abs/1907.10326

[9] R. Ranftl, V. Vineet, Q. Chen, and V. Koltun, "Dense monocular depth estimation in complex dynamic scenes," in *Proc. IEEE Conf. Comput. Vis. Pattern Recognit. (CVPR)*, Jun. 2016, pp. 4058–4066.

[10] A. Gordon, H. Li, R. Jonschkowski, and A. Angelova, "Depth from videos in the wild: Unsupervised monocular depth learning from unknown cameras," in *Proc. IEEE/CVF Int. Conf. Comput. Vis. (ICCV)*, Oct. 2019, pp. 8977–8986.

[11] Z. Li, T. Dekel, F. Cole, R. Tucker, N. Snavely, C. Liu, and W. T. Freeman, "Learning the depths of moving people by watching frozen people," in *Proc. IEEE/CVF Conf. Comput. Vis. Pattern Recognit. (CVPR)*, Jun. 2019, pp. 4521–4530.

[12] T. Van Dijk and G. De Croon, "How do neural networks see depth in single images?" in *Proc. IEEE/CVF Int. Conf. Comput. Vis. (ICCV)*, Oct. 2019, pp. 2183–2191.

[13] B. Biggio, I. Corona, D. Maiorca, B. Nelson, N. Šrndić, P. Laskov, G. Giacinto, and F. Roli, "Evasion attacks against machine learning at test time," in *Proc. Joint Eur. Conf. Mach. Learn. Knowl. Discovery Databases (ECML PKDD)*, 2013, pp. 387–402.

[14] C. Szegedy, W. Zaremba, I. Sutskever, J. Bruna, D. Erhan, I. Goodfellow, and R. Fergus, "Intriguing properties of neural networks," in *Proc. Int. Conf. Learn. Represent. (ICLR)*, 2014. [Online]. Available: https://arxiv.org/abs/1312.6199

[15] I. J. Goodfellow, J. Shlens, and C. Szegedy, "Explaining and harnessing adversarial examples," in *Int. Conf. Learn. Represent. (ICLR)*, 2015. [Online]. Available: https://arxiv.org/abs/1412.6572

[16] N. Carlini and D. Wagner, "Towards evaluating the robustness of neural networks," in *Proc. IEEE Symp. Secur. Privacy (SP)*, May 2017, pp. 39–57.

[17] S.-M. Moosavi-Dezfooli, A. Fawzi, and P. Frossard, "DeepFool: A simple and accurate method to fool deep neural networks," in *Proc. IEEE Conf. Comput. Vis. Pattern Recognit. (CVPR)*, Jun. 2016, pp. 2574–2582.

[18] N. Papernot, P. McDaniel, S. Jha, M. Fredrikson, Z. B. Celik, and A. Swami, "The limitations of deep learning in adversarial settings," in *Proc. IEEE Eur. Symp. Secur. Privacy (EuroS&P)*, Mar. 2016, pp. 372–387.

[19] T. Gittings, S. Schneider, and J. Collomosse, "Robust synthesis of adversarial visual examples using a deep image prior," in *Proc. Brit. Mach. Vis. Conf. (BMVC)*, 2019. [Online]. Available: https://bmvc2019.org/wp-content/uploads/papers/0165-paper.pdf

[20] J. H. Metzen, M. C. Kumar, T. Brox, and V. Fischer, "Universal adversarial perturbations against semantic image segmentation," in *Proc. IEEE Int. Conf. Comput. Vis. (ICCV)*, Oct. 2017, pp. 2755–2764.

[21] C. Xie, J. Wang, Z. Zhang, Y. Zhou, L. Xie, and A. Yuille, "Adversarial examples for semantic segmentation and object detection," in *Proc. IEEE Int. Conf. Comput. Vis. (ICCV)*, Oct. 2017, pp. 1369–1378.

[22] X. Wei, S. Liang, N. Chen, and X. Cao, "Transferable adversarial attacks for image and video object detection," in *Proc. 28th Int. Joint Conf. Artif. Intell.*, Aug. 2019, pp. 954–960.

[23] Z. Zhang, X. Zhu, Y. Li, X. Chen, and Y. Guo, "Adversarial attacks on monocular depth estimation," 2020, *arXiv:2003.10315*. [Online]. Available: http://arxiv.org/abs/2003.10315

[24] T. B. Brown, D. Mané, A. Roy, M. Abadi, and J. Gilmer, "Adversarial patch," 2017, *arXiv:1712.09665*. [Online]. Available: http://arxiv.org/abs/1712.09665

[25] S. Komkov and A. Petiushko, "AdvHat: Real-world adversarial attack on ArcFace face ID system," 2019, *arXiv:1908.08705*. [Online]. Available: http://arxiv.org/abs/1908.08705

[26] S. Thys, W. V. Ranst, and T. Goedeme, "Fooling automated surveillance cameras: Adversarial patches to attack person detection," in *Proc. IEEE/CVF Conf. Comput. Vis. Pattern Recognit. Workshops (CVPRW)*, Jun. 2019, pp. 1–7.

[27] A. Ranjan, J. Janai, A. Geiger, and M. Black, "Attacking optical flow," in *Proc. IEEE/CVF Int. Conf. Comput. Vis. (ICCV)*, Oct. 2019, pp. 2404–2413.

[28] A. Kurakin, I. Goodfellow, and S. Bengio, "Adversarial examples in the physical world," in *Int. Conf. Learn. Represent. (ICLR)*, 2017. [Online]. Available: https://arxiv.org/abs/1607.02533

[29] A. Athalye, L. Engstrom, A. Ilyas, and K. Kwok, "Synthesizing robust adversarial examples," in *Proc. Int. Conf. Mach. Learn. (ICML)*, 2018, pp. 284–293.

[30] K. Eykholt, I. Evtimov, E. Fernandes, B. Li, A. Rahmati, C. Xiao, A. Prakash, T. Kohno, and D. Song, "Robust physical-world attacks on deep learning visual classification," in *Proc. IEEE/CVF Conf. Comput. Vis. Pattern Recognit.*, Jun. 2018, pp. 1625–1634.

[31] J. Lu, H. Sibai, E. Fabry, and D. Forsyth, "NO need to worry about adversarial examples in object detection in autonomous vehicles," 2017, *arXiv:1707.03501*. [Online]. Available: http://arxiv.org/abs/1707.03501

[32] D. Eigen, C. Puhrsch, and R. Fergus, "Depth map prediction from a single image using a multi-scale deep network," in *Proc. Adv. Neural Inf. Process. Syst. (NIPS)*, 2014, pp. 2366–2374.

[33] A. Saxena, S. H. Chung, and A. Y. Ng, "Learning depth from single monocular images," in *Proc. Adv. Neural Inf. Process. Syst. (NIPS)*, 2006, pp. 1161–1168.

[34] A. Saxena, M. Sun, and A. Y. Ng, "Make3D: Learning 3D scene structure from a single still image," *IEEE Trans. Pattern Anal. Mach. Intell.*, vol. 31, no. 5, pp. 824–840, May 2009.

[35] F. Liu, C. Shen, G. Lin, and I. Reid, "Learning depth from single monocular images using deep convolutional neural fields," *IEEE Trans. Pattern Anal. Mach. Intell.*, vol. 38, no. 10, pp. 2024–2039, Oct. 2016.

[36] X. Wang, D. F. Fouhey, and A. Gupta, "Designing deep networks for surface normal estimation," in *Proc. IEEE Conf. Comput. Vis. Pattern Recognit. (CVPR)*, Jun. 2015, pp. 539–547.

[37] L. Ladicky, J. Shi, and M. Pollefeys, "Pulling things out of perspective," in *Proc. IEEE Conf. Comput. Vis. Pattern Recognit.*, Jun. 2014, pp. 89–96.

[38] K. Karsch, C. Liu, and S. B. Kang, "Depth transfer: Depth extraction from video using non-parametric sampling," *IEEE Trans. Pattern Anal. Mach. Intell.*, vol. 36, no. 11, pp. 2144–2158, Nov. 2014.

[39] B. Li, C. Shen, Y. Dai, A. van den Hengel, and M. He, "Depth and surface normal estimation from monocular images using regression on deep features and hierarchical CRFs," in *Proc. IEEE Conf. Comput. Vis. Pattern Recognit. (CVPR)*, Jun. 2015, pp. 1119–1127.

[40] Y. Cao, Z. Wu, and C. Shen, "Estimating depth from monocular images as classification using deep fully convolutional residual networks," *IEEE Trans. Circuits Syst. Video Technol.*, vol. 28, no. 11, pp. 3174–3182, Nov. 2018.

[41] W. Chen, Z. Fu, D. Yang, and J. Deng, "Single-image depth perception in the wild," in *Proc. Adv. Neural Inf. Process. Syst. (NIPS)*, 2016, pp. 730–738.

[42] Y. Kuznietsov, J. Stuckler, and B. Leibe, "Semi-supervised deep learning for monocular depth map prediction," in *Proc. IEEE Conf. Comput. Vis. Pattern Recognit. (CVPR)*, Jul. 2017, pp. 6647–6655.

[43] J. Xie, R. Girshick, and A. Farhadi, "Deep3D: Fully automatic 2D-to-3D video conversion with deep convolutional neural networks," in *Proc. Eur. Conf. Comput. Vis. (ECCV)*, 2016, pp. 842–857.

[44] C. Fehn, "Depth-image-based rendering (DIBR), compression, and transmission for a new approach on 3D-TV," *Proc. SPIE*, vol. 5291, pp. 93–104, May 2004.

[45] R. Garg, V. K. BG, G. Carneiro, and I. Reid, "Unsupervised cnn for single view depth estimation: Geometry to the rescue," in *Proc. Eur. Conf. Comput. Vis. (ECCV)*, 2016, pp. 740–756.

[46] A. Dosovitskiy, P. Fischer, E. Ilg, P. Hausser, C. Hazirbas, V. Golkov, P. V. D. Smagt, D. Cremers, and T. Brox, "FlowNet: Learning optical flow with convolutional networks," in *Proc. IEEE Int. Conf. Comput. Vis. (ICCV)*, Dec. 2015, pp. 2758–2766.

[47] M. Jaderberg, K. Simonyan, and A. Zisserman, "Spatial transformer networks," in *Proc. Adv. Neural Inf. Process. Syst. (NIPS)*, 2015, pp. 2017–2025.

[48] K. Simonyan and A. Zisserman, "Very deep convolutional networks for large-scale image recognition," in *Proc. Int. Conf. Learn. Represent. (ICLR)*, 2015. [Online]. Available: https://arxiv.org/abs/1409.1556

[49] G. Hinton, O. Vinyals, and J. Dean, "Distilling the knowledge in a neural network," *Stat*, vol. 1050, p. 9, 2015.

[50] D. P. Kingma and J. Ba, "Adam: A method for stochastic optimization," in *Proc. Int. Conf. Learn. Represent. (ICLR)*, 2015. [Online]. Available: https://arxiv.org/abs/1412.6980

[51] M. Sharif, S. Bhagavatula, L. Bauer, and M. K. Reiter, "Accessorize to a crime: Real and stealthy attacks on state-of-the-art face recognition," in *Proc. ACM SIGSAC Conf. Comput. Commun. Secur. (ACM SIGSAC)*, 2016, pp. 1528–1540.

[52] A. Mahendran and A. Vedaldi, "Understanding deep image representations by inverting them," in *Proc. IEEE Conf. Comput. Vis. Pattern Recognit. (CVPR)*, Jun. 2015, pp. 5188–5196.

[53] G. Huang, Z. Liu, L. Van Der Maaten, and K. Q. Weinberger, "Densely connected convolutional networks," in *Proc. IEEE Conf. Comput. Vis. Pattern Recognit. (CVPR)*, Jul. 2017, pp. 4700–4708.

[54] L.-C. Chen, G. Papandreou, I. Kokkinos, K. Murphy, and A. L. Yuille, "DeepLab: Semantic image segmentation with deep convolutional nets, atrous convolution, and fully connected CRFs," *IEEE Trans. Pattern Anal. Mach. Intell.*, vol. 40, no. 4, pp. 834–848, Apr. 2018.

[55] J. Uhrig, N. Schneider, L. Schneider, U. Franke, T. Brox, and A. Geiger, "Sparsity invariant CNNs," in *Proc. Int. Conf. 3D Vis. (3DV)*, Oct. 2017, pp. 11–20.

[56] A. Geiger, P. Lenz, C. Stiller, and R. Urtasun, "Vision meets robotics: The KITTI dataset," *Int. J. Robot. Res.*, vol. 32, no. 11, pp. 1231–1237, Sep. 2013.

[57] A. Paszke, S. Gross, F. Massa, A. Lerer, J. Bradbury, G. Chanan, T. Killeen, Z. Lin, N. Gimelshein, and L. Antiga, "PyTorch: An imperative style, high-performance deep learning library," in *Proc. Adv. Neural Inf. Process. Syst. (NIPS)*, 2019, pp. 8024–8035.

[58] R. R. Selvaraju, M. Cogswell, A. Das, R. Vedantam, D. Parikh, and D. Batra, "Grad-CAM: Visual explanations from deep networks via gradient-based localization," in *Proc. IEEE Int. Conf. Comput. Vis. (ICCV)*, Oct. 2017, pp. 618–626.

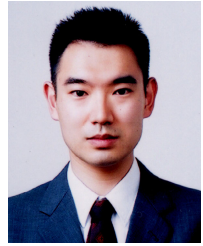

**RYUTAROH MATSUMOTO** (Member, IEEE) was born in Nagoya, Japan, in November 29, 1973. He received the B.E. degree in computer science, the M.E. degree in information processing, and the Ph.D. degree in electrical and electronic engineering from the Tokyo Institute of Technology, Japan, in 1996, 1998, and 2001, respectively. He was an Assistant Professor, from 2001 to 2004, and an Associate Professor, from 2004 to 2017 with the Department of Information and Communications Engineering, Tokyo Institute of Technology. He also served as a Velux Visiting Professor with the Department of Mathematical Sciences, Aalborg University, Denmark, in 2011 and 2014. From 2017 to 2020, he was an Associate Professor with the Department of Information and Communication Engineering, Nagoya University, Japan, under the 6-University Human Assets Promotion Program for Innovative Education and Research Program (6-U HAPPIER). He returned to the Tokyo Institute of Technology after the expiration of the 6-U HAPPIER. His research interests include error-correcting codes, quantum information theory, information theoretic security, and communication theory. He received the Young Engineer Award from IEICE and the Ericsson Young Scientist Award from Ericsson Japan, in 2001. He received the best paper awards from IEICE in 2001, 2008, 2011, and 2014.

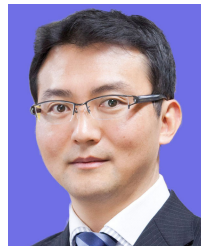

**KEITA TAKAHASHI** (Member, IEEE) received the B.E., M.S., and Ph.D. degrees in information and communication engineering from The University of Tokyo, in 2001, 2003, and 2006, respectively. He was a Project Assistant Professor with The University of Tokyo, from 2006 to 2011, and an Assistant Professor with The University of Electro-Communications, from 2011 to 2013. He is currently an Associate Professor with the Graduate School of Engineering, Nagoya University, Japan. His research interests include computational photography, image-based rendering, and 3-D displays.

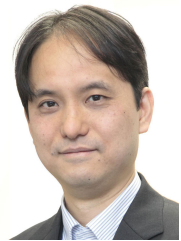

**TOSHIAKI FUJII** (Member, IEEE) received the B.E., M.E., and Dr.E. degrees in electrical engineering from The University of Tokyo, in 1990, 1992, and 1995, respectively. Since 1995, he has been with the Graduate School of Engineering, Nagoya University, where he is currently a Professor. From 2008 to 2010, he was with the Graduate School of Science and Engineering, Tokyo Institute of Technology. His research interests include multidimensional signal processing, multicamera systems, multiview video coding and transmission, free-viewpoint television, and related applications.

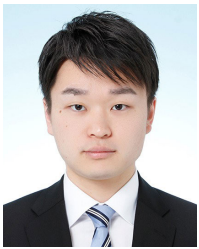

**KOICHIRO YAMANAKA** received the B.E. degree in electrical engineering from Nagoya University, Japan, in 2019, where he is currently a Graduate Student with the Graduate School of Engineering. His research interests include monocular depth estimation and adversarial examples.

•••